\documentclass{ws-procs11x85}
\usepackage{ws-procs-thm}           
\usepackage{bbm}
\usepackage{algorithm}
\usepackage[noend]{algpseudocode}
\usepackage{multirow,xcolor}

\def\reals{\mathbb{R}}

\newcommand{\anie}[1]{{\color{black}#1}}

\begin{document}

\title{LitGen: Genetic Literature Recommendation Guided by Human Explanations}

\author{Allen Nie$^{1,4}$, Arturo L. Pineda$^1$, Matt W. Wright$^{1,2}$, Hannah Wand$^{1,2,3}$, Bryan Wulf$^1$, Helio A. Costa$^1,2$, Ronak Y. Patel$^5$, Carlos D. Bustamante$^{1,6}$, James Zou$^{1,4,6}$}

\address{$^1$Department of Biomedical Data Science, Stanford University School of Medicine
\\
$^2$Department of Pathology, Stanford University School of Medicine\\
$^3$Department of Cardiology, Stanford Healthcare \\
$^4$Department of Computer Science, Stanford University \\
$^5$Department of Molecular and Human Genetics, Baylor College of Medicine \\
$^6$Chan-Zuckerberg Biohub
}







\begin{abstract}
As genetic sequencing costs decrease, the lack of clinical interpretation of  variants has become the bottleneck in using genetics data. A major rate limiting step in clinical interpretation is the manual curation of evidence in the genetic literature by highly trained biocurators. 
What makes curation particularly time-consuming is that the curator needs to identify papers that study variant pathogenicity using different types of approaches and evidences---e.g. biochemical assays or case control analysis.
In collaboration with the Clinical Genomic Resource (ClinGen)---the flagship NIH program for clinical curation---we propose the first machine learning system, LitGen, that can retrieve papers for a particular variant and filter them by specific evidence types used by curators to assess for pathogenicity. LitGen uses semi-supervised deep learning to predict the type of evi+dence provided by each paper. It is trained on papers annotated by ClinGen curators and systematically evaluated on new test data collected by ClinGen. 
LitGen further leverages rich human explanations and unlabeled data to gain 7.9\%-12.6\% relative performance improvement over models learned only on the annotated papers.
It is a useful framework to improve clinical variant curation.  
\end{abstract}


\copyrightinfo{\copyright\ 2019 The Authors. Open Access chapter published by World Scientific Publishing Company and distributed under the terms of the Creative Commons Attribution Non-Commercial (CC BY-NC) 4.0 License.}


\section{Introduction}

The diversity of genetic variations that exist in the modern human population are slowly been recognized and discovered. 
Some of these variations are responsible for well-known physical differentiation across humans (e.g. hair color \cite{branicki2011model}), other variants can predict the development of inherited diseases like sickle-cell anemia or cystic fibrosis, and a few others are protective of disease, like some variations of PCSK9 which lowers the risk for coronary heart disease \cite{cohen2006sequence}. However, little is known overall about the more than 650 million variants known to date across the human genome \cite{pawliczek2018clingen}. In PubMed using the search term ‘genetic variation’ returns over one million manuscripts, with almost half of them generated in the last 10 years. 

Our understanding of previous published studies linking human genetic variants with medical syndromes and phenotypic traits is still limited. 
In 2013, the United States’ National Center for Biotechnology Information (NCBI) established the Clinical Genome
 program~\cite{rehm2015clingen}, with the goal of defining the clinical relevance of key genes and variants through several gene and variant curation expert panels. These experts meet regularly to consider new evidence in the literature to curate and assess the pathogenicity of variants. The variant curation process combines clinical, genetic, population, and functional evidence with expert review to classify variants into 1 of 5 categories (Pathogenic, Likely Pathogenic, Variant of Unknown Significance, Likely Benign, Benign) according to the joint 2015 American College of Medical Genetics (ACMG), and Association for Medical Pathology (AMP) guidelines on clinical significance \cite{richards2015standards}.

The ACMG/AMP guidelines provide a set of criteria, and a curator searches for evidence and evaluates whether or not the evidence is sufficient to mark each criterion as met. A pathogenicity classification for each variant is calculated from the totality of the evidence evaluated using the ACMG/AMP criteria. Many of these criteria are mostly evaluated using pertinent information gleaned from publications, and finding the relevant publications that contain relevant evidence is a significant challenge to curators. 


\smallskip
\noindent \textbf{The workflow of curating variants of clinical relevance.}
The ClinGen procedure for biocurators\footnote{https://clinicalgenome.org/site/assets/files/3677/clingen\_variant-curation\_sopv1.pdf} defines four steps to assess the pathogenicity of a variant: 1) select a variant of interest with and the suspected disease or mode of inheritance; 2) review available literature evidence about the disease; 3) curate evidence according to the ACMG/AMP criteria; 4) propose a level of pathogenicity. This process is assisted by ClinGen’s Variant Curation Interface\footnote{https://clinicalgenome.org/curation-activities/variant-pathogenicity/}. Biocurators outside of the ClinGen environment follow a similar procedure. 
In the third step,  when biocurators consider each of the ACMG/AMP criteria to systematically evaluate if the considered variant has some available literature. VCI further groups ACMG/AMP criteria into evidence types, many of which require evidence from published literature. Assessing which paper is relevant for each of the evidence types has a high burden of time and effort on the biocurator. To the best of our knowledge, there is currently no tool to automatically facilitate this task.

\smallskip
\noindent \textbf{Our contribution}
We built a machine learning system LitGen that recommends papers to biocurators based on the evidence types presented in the paper.
We believe this is the first system that analyzes papers for content on clinically relevant evidence types beyond variant name normalization or information matching~\cite{wei2017tmvar,Birgmeier171322,kuleshov2019machine}. We also contribute to the research area of  semi-supervised learning with explanations. LitGen effectively uses explanations to guide semi-supervised learning. A thorough evaluation on new ClinGen data demonstrates that LitGen outperforms competitive baselines by a large margin.





\section{Clinical Variant Curation Data}


\subsection{ClinGen's Variant Curation Interface (VCI)}


The data that we use to develop LitGen are collected through ClinGen’s Variant Curation Interface (VCI). VCI is a curation web tool that was designed to support variant curation based on the ACMG/AMP Guidelines and serves as a platform for the standardized curation of clinical variants by ClinGen’s Variant Curation Expert Panels. This pool of evidence can then be utilized by all VCI users when evaluating each of the ACMG/AMP criteria in turn within the interface. The VCI allows a user to provide an explanation comment describing the rationale for their evaluation in a text field, and to provide a PubMed ID linking to the relevant published literature that contains the data that supports their evaluation. The VCI allows the curator to assert whether the paper is relevant for a subset of evidence types. Here we focus on the five most common evidence types (Fig. 1).



\begin{figure}[h]
\centerline{
\includegraphics[scale=0.6]{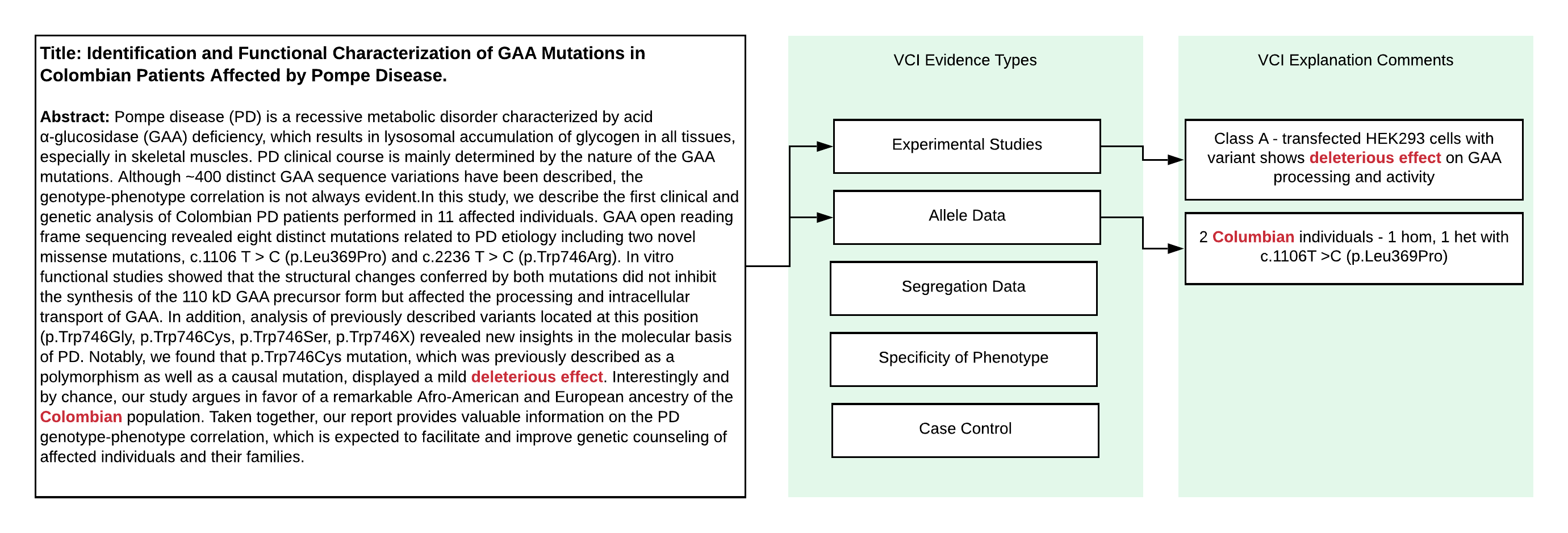} 
}
\caption{Paper annotation workflow. From a paper on PubMed (left), the curator selects which subset of the five variant curation (VCI)  evidence types that the paper is relevant for (middle), and provide explanations for the selection (right). We highlight some keywords for emphasis. LitGen's goal is to predict which evidence types are relevant  given a paper.}
\label{fig:vci-screenshot}
\end{figure}

\vspace{-0.3cm}

\subsection{Labeled papers} 


We extracted all papers entered by VCI users between October 2016 and March 2019. The collected data include 1543 unique papers which contained clinical information on 932 unique variants. We randomly split this set of papers into train, dev, and test set by 0.9/0.05/0.05. Additionally, we collected a new set of 358 papers entered from April 2019 to May 2019 as a holdout evaluation dataset. Papers in this holdout evaluation dataset are entirely new. Table ~\ref{tab:labeled-data} shows the distribution of these two datasets. Each paper contains information that can be categorized into different evidence types that curators used to assert clinical pathogenicity. Curators can optionally provide an explanation comment for each type of evidence. In this manuscript, we focused on the top 5 VCI evidence types by the number of unique papers---these are Case Control, Specificity of Phenotype, Allele Data, Experimental Studies, and Segregation Data. These 5 evidence types covers 84\% of all papers annotated in the VCI. 

\begin{table}[h]
	\footnotesize
	\tbl{Labeled data summary: number of papers and explanations by VCI evidence type.}
	{\begin{tabular}{@{}lccccc@{}}
			\toprule
			&  & \multicolumn{2}{c}{\# unique papers} & \multicolumn{2}{c}{\# explanations}  \\ 
			Evidence types in the VCI & ACMG criteria & Train & Holdout & Train & Holdout \\ \colrule
			Experimental Studies & BS3, PS3 & 385 & 74 & 732 & 80  \\ \colrule 
			Allele Data & BP2, PM3 & 441 & 86 & 971 & 103  \\ \colrule 
			Segregation Data & BS4, PP1 & 232 & 40 & 271 & 40 \\ \colrule 
			Specificity of Phenotype & PP4 & 482 & 26 & 993 & 28 \\ \colrule 
			Case Control & PS4 & 656 & 264 & 952 & 331 \\ \colrule
			\textbf{Total} & & \textbf{1543} & \textbf{358} & \textbf{3919} & \textbf{582} \\
            \botrule
	\end{tabular}}
	\small
	Training data collected during Oct 2016 to Mar 2019. Holdout evaluation data collected during April 2019 to May 2019. Note that we do not allow the algorithm to use explanations during test time. We have 1543 labeled data points for training.
	\label{tab:labeled-data}
\end{table}



\subsection{Unlabeled papers}



In order to investigate whether semi-supervised learning can improve our model's performance, we collect a larger set of unlabeled papers through the following pipeline. We use ClinGen Allele Registry~\cite{pawliczek2018clingen} to find the rsid of the variant if a clinical variant ID is provided. We use LitVar API, a new service provided by NCBI~\cite{allot2018litvar}, to retrieve relevant literature of a given variant. LitVar scanned and indexed all of PubMed abstracts and PubMed Central full papers. 
We use this pipeline to retrieve all relevant papers to all variants curated through the ClinGen VCI. ClinGen Allele Registry found rsid for 877 of 932 variants (94.1\%). We further found 742 (79.6\%) variants that have been mentioned in the literature indexed in LitVar. We queried 4477 papers in total from LitVar, and 650 of these papers overlap with papers that have already entered into ClinGen by curators. Excluding these papers, we have 3827 new papers.
We release all of our code and data at \url{https://github.com/windweller/ClinGenML/}.



\section{Method}


We use the following notations to describe our data. 
Each paper in our dataset is annotated with at least one VCI evidence type and the associated explanation comments on the rationale of selection.
For the labeled papers dataset, we have $(x, \bm{y}) \in (\mathcal{X}, \bm{\mathcal{Y}})$ where $\bm{y} \in [0, 1]^m$ for m labels and $ m = 5$ in our case. Here $x$ represents the paper title and abstract. This is a multi-label setting because each paper can contain multiple evidence types. Each explanation comment is associated with \anie{exactly} one evidence type. We can regard it as $(e, y) \in (E, \mathcal{Y})$, where $e$ is the explanation text and $y \in \{1,...,m\}$ describes the evidence type. 



\subsection{BiLSTM baseline}

We aim to train a competitive supervised learning algorithm on the labeled data.
We use the state-of-the-art text processing algorithm for our model: long-short-term memory networks (LSTMs). It has been used in many natural language processing applications~\cite{mikolov2012statistical}, generating complex human responses~\cite{nie2019learning}, and well-adopted in clinical text processing~\cite{nie2018deeptag,zhang2019vettag}. We use the bidirectional variant of this algorithm proposed by Graves et al.~\cite{graves2005bidirectional}. 

For each paper abstract $x = w_1, ..., w_T$, we compute the hidden state vectors $H = [h_1, ..., h_T]$. We compute the vector representation of the abstract $c(x)$ using the global max-pooling over the temporal dimension suggested by Collobert \& Weston \cite{collobert2008unified}. At last, we predict whether an evidence type $y_i$ exist through a sigmoid binary classifier with parameter $\theta_i$. We compute the binary cross-entropy loss through the predicted labels $\bm{\hat{y}} = [\hat y_1, ..., \hat y_m]$ and true labels $\bm{y}$.

\begin{eqnarray}
&H = [h_1, ..., h_T] = \text{BiLSTM}(w_1,...,w_T) \text{, } H \in \reals^{T \times d} \\
&\bm{c}(x) = [\max(H_{\cdot, 1}), \max(H_{\cdot, 2}), ..., \max(H_{\cdot, d})], \bm{c}(x) \in \reals^{d} \\
&P(y_i) = \hat{y}_i = \sigma(\theta_i^\intercal \bm{c}(x)) \text{, for } i=1,...,m \\
& \mathcal{L}_{\text{BCE}}(x, \bm{\hat{y}}, \bm{y}) = - \frac{1}{m} \sum_{i=1}^m y_i \log (\hat y_i) + (1 - y_i)\log(1- \hat y_i)
\end{eqnarray}

\subsection{Leveraging unlabeled data}


After training a competitive baseline model only on limited labeled data, we explore the possibility of leveraging unlabeled paper by using a proxy labeling model. Proxy-label approach to semi-supervised learning has been generally shown to improve the performance of the final model. This approach aims to produce proxy labels on unlabeled data, which later are used as targets together with labeled data to train the final model. These proxy labels do not reflect the ground truth labels, but they might provide some signals for learning~\cite{blum1998combining,zhou2004democratic}. 



\begin{figure}[h]
\centerline{
\includegraphics[scale=0.5]{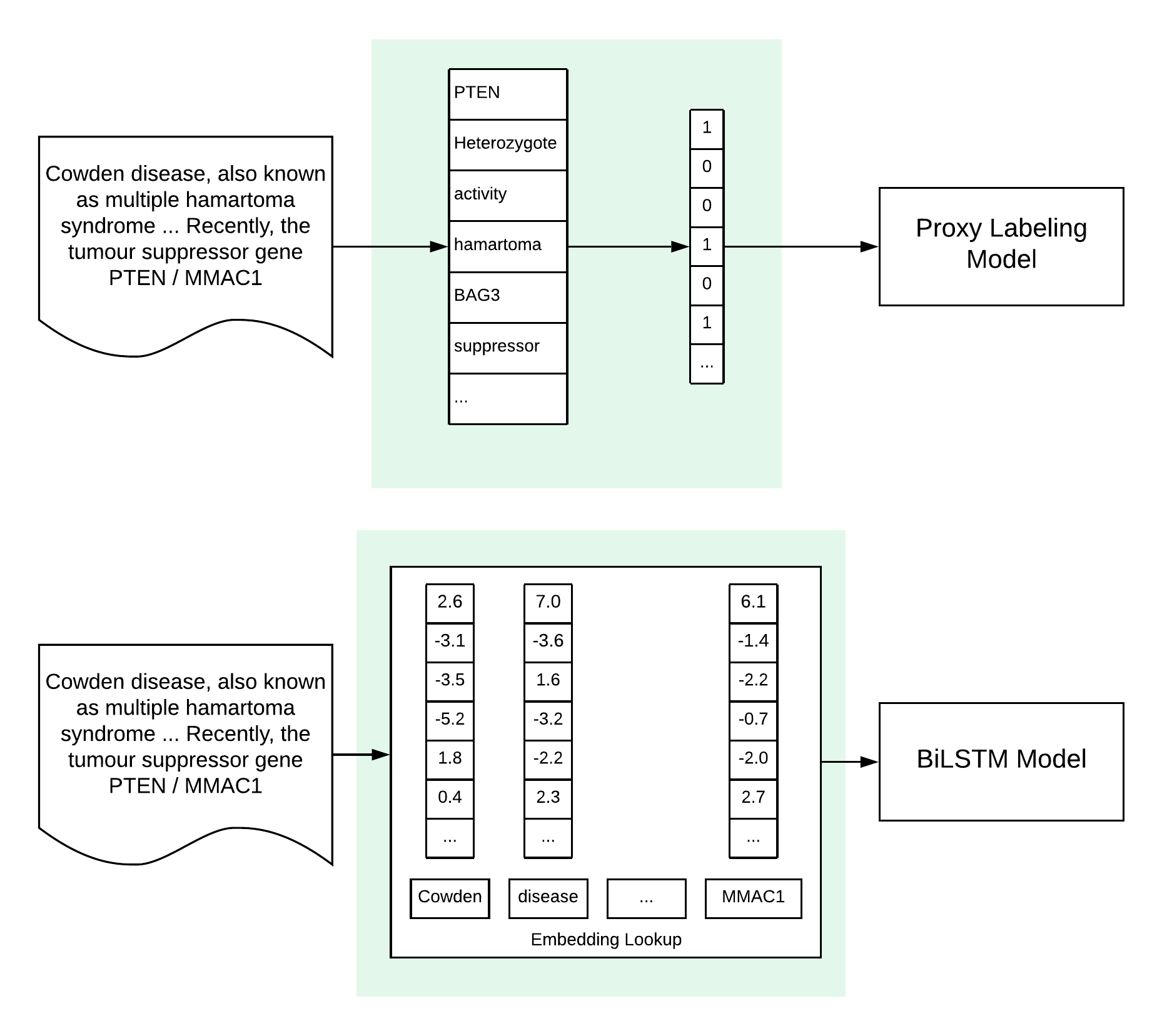} 
}
\caption{\textbf{Naive Unlabeled:} The two views of inputs for the proxy labeling model and the BiLSTM model.}
\label{fig:input-pipeline}
\end{figure}


We train a random forest model to predict evidence types on our labeled dataset (represented as bag-of-words). We then apply this random forest to predict labels for each unlabeled paper; we call these the proxy labels. Finally we train our BiLSTM model on proxy-labeled unlabeled data and labeled data together. We refer to this strategy as \textbf{Naive Unlabeled}, because \anie{it is} a simple and direct approach to use the unlabeled papers. The point of using the random forest to generate the proxy labels is that it contains different inductive bias compared to the original \anie{BiLSTM}. 
Zhou \& Goldman~\cite{zhou2004democratic} showed that when the proxy labeling models have different bias compared to the final classifier, the generated proxy labels can often improve the model's performance.




\subsection{Explanations in multitask learning}



Beyond building a strong BiLSTM baseline and incorporate proxy labeling methods on unlabeled data, another important feature of our curation dataset is that we have human-provided explanations associated with each paper. Each explanation is a concise summary of \emph{why} the curator asserted that a paper provides a particular type of pathogenicity evidence. We hypothesized that these explanations could help us to generating features that are salient for evidence predictions. 
Contrary to using humans to label each training example, which is very costly both in terms of time and resource, recent works have  explored whether human-provided explanations will allow models to learn beyond instance-level labels. Early works focus on using semantic parsing over human explanations to obtain labeling functions~\cite{srivastava2017joint,hancock2018training}. However, such approaches are limited to explanations that have fixed format such as ``X because of Y and Z''. The explanations provided by our curators are free text and do not conform to predefined templates. An innovation of our work is on how to leverage these explanations. 

\begin{figure}[h]
\centerline{
\includegraphics[scale=0.4]{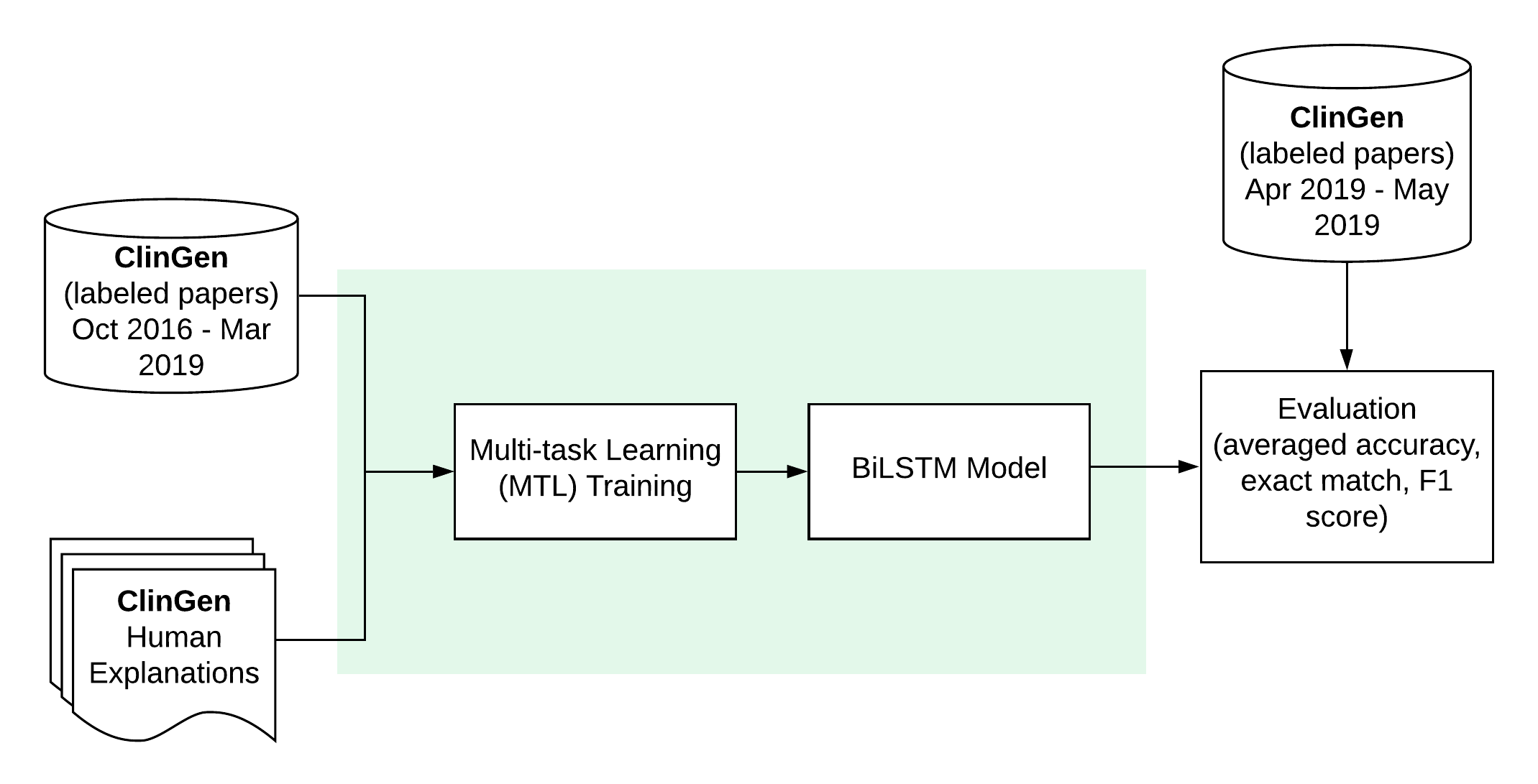} 
}
\caption{Multi-task learning pipeline that leverages labeled data and explanations}
\label{fig:mtl-pipeline}
\end{figure}

A simple way to use the explanations is to treat them simply as additional labeled examples where the label is the associated VCI evidence type.
 We build a multi-task learning objective, where the BiLSTM model is asked to optimize for two tasks: predicting whether a paper contains information relevant to a VCI evidence type (original task, loss marked as $\mathcal{L}_1$), as well as whether an explanation is provided as rationale for a VCI evidence type (explanation prediction task, loss marked as $\mathcal{L}_2$). For each epoch, we train on two tasks separately: first on the explanation prediction task, and after iterating through all batches of explanations, we train on the original paper abstract prediction task. We use a scalar hyperparameter $\lambda \in [0, 1]$ to scale the loss of the explanation prediction task. We call this approach \textbf{Naive exp}.



There are inherent problems to this approach. First of all, when we train on $(e, y)$, the explanations have a different length distribution compares to $x$, the paper abstracts. Explanations tend to be shorter and more succinct. Since we are using the same BiLSTM model to process both texts, we are learning from two data distributions. Second, even though both explanations and papers are associated with a VCI evidence type, one explanation can only exclusively be used to justify for one VCI evidence type, while a paper can be associated with multiple VCI evidence types. Therefore, the nature of data-to-label mapping is different for the two tasks. The last problem is that explanations are noisy. Curators submit these explanations often as a comment or additional information to support their choice of paper. Not all words in explanations are useful for the original task. We address all three problems by proposing our new approach: use explanations to perform feature selection, and then use the selected features to proxy label the unlabeled papers. 

\subsection{Explanations as feature selection for proxy labeling}


\begin{figure}[h]
\centerline{
\includegraphics[scale=0.4]{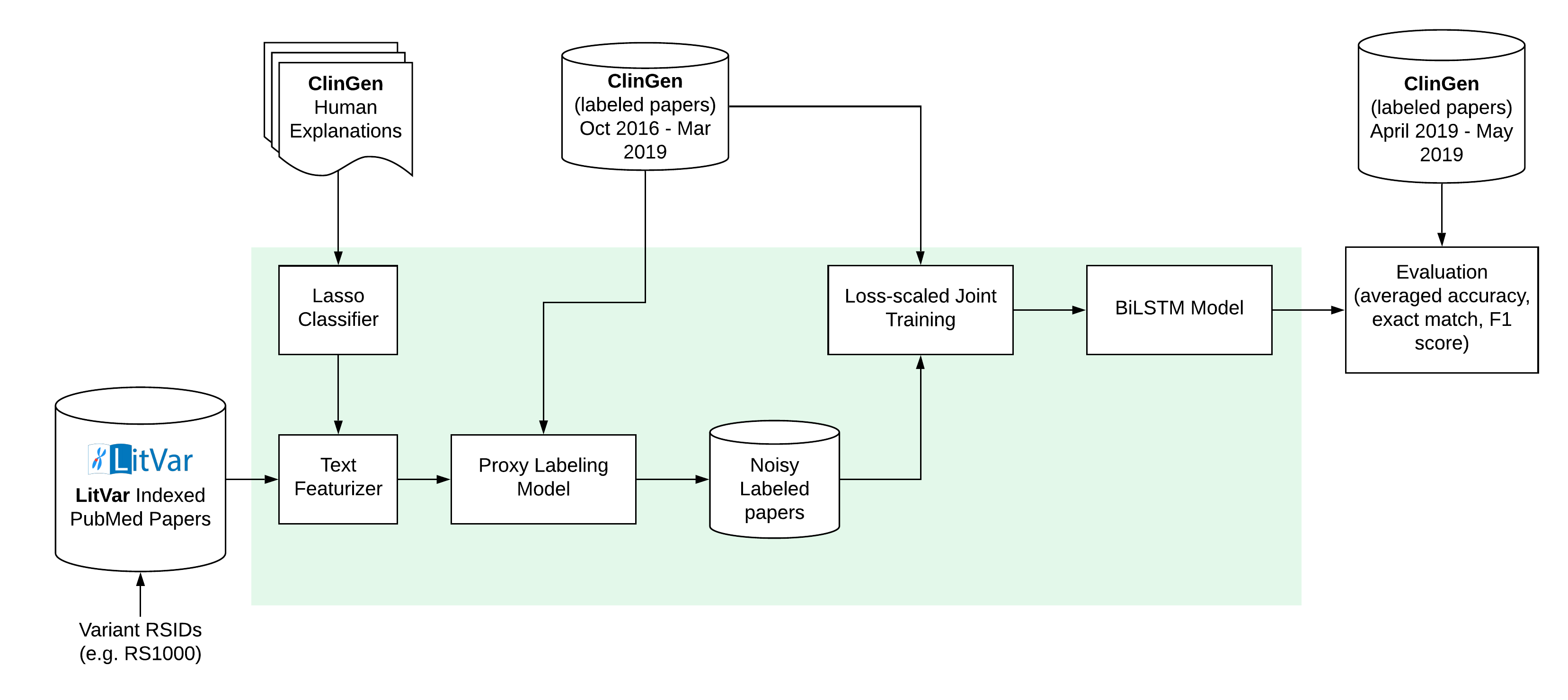} 
}
\caption{General training pipeline that leverages unlabeled data guided by explanations.}
\label{fig:model-pipeline}
\end{figure}

\begin{figure}[h]
\centerline{
\includegraphics[scale=0.7]{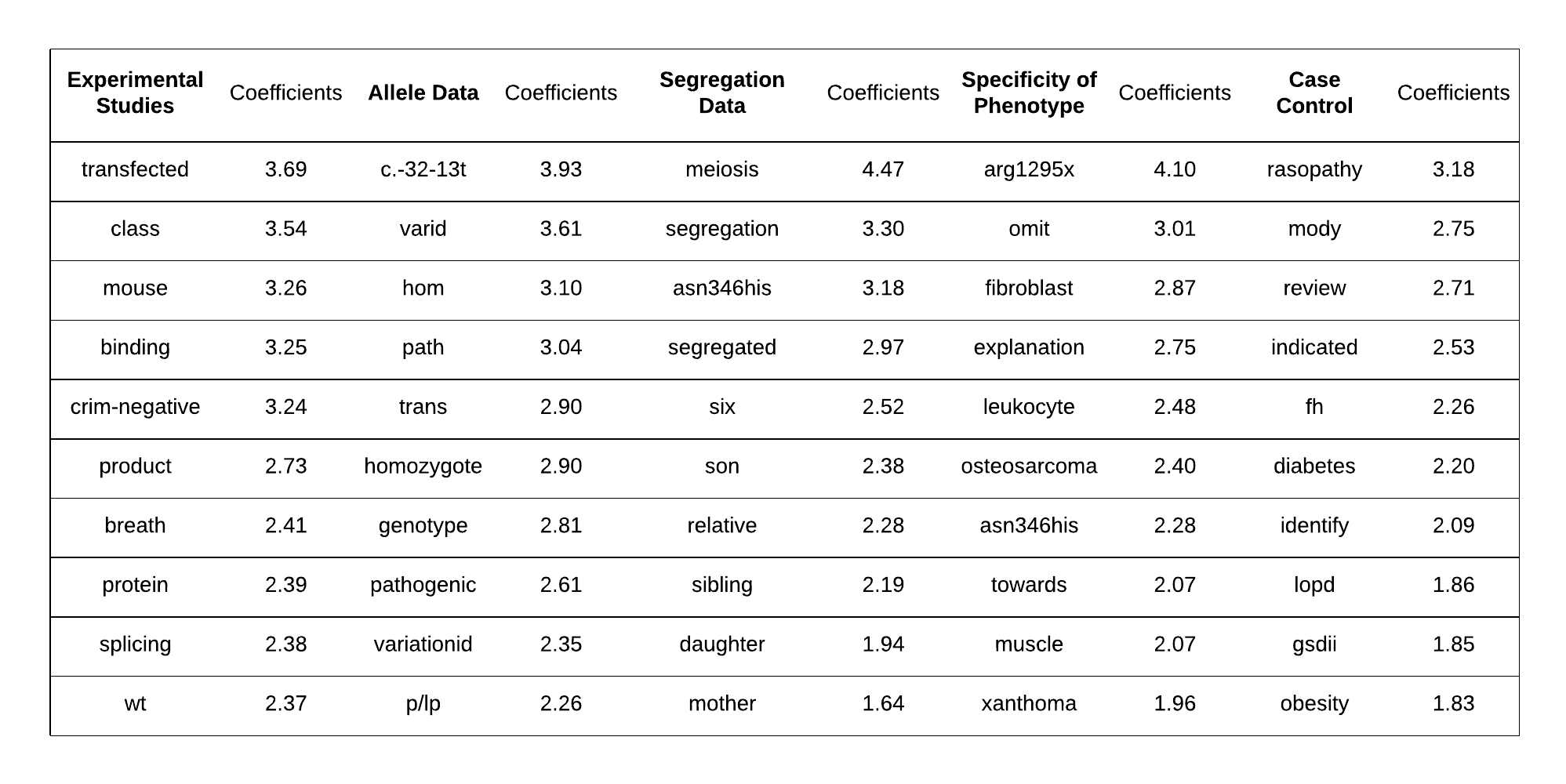} 
}
\caption{We \anie{display} a set of keywords that are the most positively associated with each VCI evidence types from human explanations by training a lasso model on unigram features. Coefficients refer to Lasso coefficients.}
\label{tab:keywords}
\end{figure}

We first train a Lasso classifier (a logistic regression classifier with $L_1$ regularization) on the frequency-encoded \anie{unigram} feature representation of the explanations. Our Lasso classifier obtains a coefficient on each word that determines whether the word is important for the prediction of which VCI evidence type an explanation is associated with. \anie{Our Lasso classifier obtains 89.0\% accuracy on this classification task}. This shows that explanations are easier to classify compared to paper abstracts and they contain useful signals that can be leveraged. We extract words that have non-zero coefficients. We display some of these words in Figure~\ref{tab:keywords}. In total, we are able to find 799 words that have non-zero coefficient out of 7550 words contained in explanations. We use these 799 words as the selected features and then follow the same proxy labeling strategy as the Naive Unlabeled algorithm. 

For each paper abstract, we record the frequency of these 799 unique words in the abstract and ignore all other words. Originally, in section 3.2, we naively encode the paper abstract obtaining a vector equivalent to the size of the entire vocabulary space (after removing stop words and punctuation), which is 16860. We have now restricted the dimension of the vector representing the paper abstract from 16860 to 799. We refer to this feature selection process as \textbf{Exp-guided}.
We then train our random forest proxy labeling model on the paper abstracts and use it to generate proxy labels for unlabeled data. At last, we train the BiLSTM model on both proxy labeled unlabeled data and ground-truth data. We refer to this setting as the \textbf{LitGen} model. 


Another advantage to our explanation-guided feature selection process is that we can now automatically generate labeling functions without semantic parsing. We use a simple heuristic to binarize the coefficients in our Lasso classifier: each of the 799 words is a labeling function. If the word has a positive coefficient, we output \anie{+1} when we encounter this word. If the word has a negative coefficient, we output -1. When the word is missing, \anie{we output 0}. This allows us to leverage labeling function aggregation algorithm such as Snorkel-MeTal~\cite{ratner2018snorkel}. We include this result to show that by selecting features from explanations, we are able to leverage multiple approaches in semi/weakly supervised learning. We refer to this setting as \textbf{Exp-guided Snorkel}.
\section{Experimental results}

\subsection{Evaluation metrics}

We use the following metrics to evaluate model performance in predicting the evidence types given a paper. We compute the average accuracy (Avg Accu) across VCI evidence types. Accuracy reflects how correctly the model can determine whether the paper contains a type of evidence or not. We compute the exact match ratio (EM) as well, which is a more strict metric that requires the model's predictions to exactly match every ground truth label. Finally, we also compute the average $F_1$ score weighted by the number of examples in each evidence type (Wgt F1). All the models are trained on data up to March 2019 and are evaluated on new ClinGen paper annotated from April to May 2019. 

\subsection{Performance comparison}

We train the \textbf{LitGen model} based on the strategies described in the method section. We also consider a baseline classifier that randomly predicts the value of each label based on the class balance of the training data. We evaluate all trained models on a final holdout set of 347 disjoint papers. In Table~\ref{tab:main-result}, we show the performance of our proposed methods to incorporate explanations into the supervised learning and proxy-label semi-supervised learning pipelines. 




%
\begin{table}[h]
	\tbl{Performance of different training strategies for  LitGen model. 
	}
	{\begin{tabular}{@{}lccc@{}} 
			\toprule
			&  \multicolumn{3}{c}{Apr 2019 to May 2019} \\ 
			Strategy & Avg Accu & EM & Wgt $F_1$ \\ \colrule
			Baseline (Majority) & 62.9 & 8.7 & 36.0 \\ \colrule
			BiLSTM & 82.6 & 45.2 & 62.7 \\
			BiLSTM + Naive Exp & 83.8 & 48.7 & 66.5 \\
			BiLSTM + Naive Unlabled & 83.9 & 50.1 & 65.7 \\
			BiLSTM + Naive Exp + Naive Unlabeled & \textit{82.9} & \textit{48.4} & \textit{66.4} \\ \colrule 
			BiLSTM + Exp-guided Snorkel & 84.0 & 50.1 & 66.8 \\   
			\textbf{LitGen}: BiLSTM + Exp-guided Unlabeled & \textbf{85.0} & \textbf{51.6} & \textbf{68.1}  \\  \botrule
	\end{tabular}}\label{tab:main-result}
\end{table}


\paragraph{Unlabeled data and explanations both help} 
We observe the improvement over BiLSTM model when training on proxy-labeled paper abstracts and leveraging explanations: both \textit{BiLSTM + Naive Unlabled} and \textit{BiLSTM + Naive Exp} outperform \textit{BiLSTM} on all the evaluation metrics. That naive training on explanations leads to improvement shows that explanations do provide learning signals for the model.

\smallskip

\paragraph{Naive joint training hurts performance} 
However, even though training on explanation prediction task or training on proxy-labeled paper abstracts each improves the final model's performance, such effect is not additive when we train on both. \emph{BiLSTM + Naive Exp + Naive Unlabeled} performs relatively poorly. We have discussed potential drawback of training naively on explanations such as text length distribution mismatch and noisy explanation text.

\smallskip
\paragraph{Using explanations for feature selection outperforms all}
Explanations contain valuable learning signals but are noisy in its writing. When we use them for feature selection, choosing words that are determined important by a Lasso classifier, we accomplish two goals at once: 1) reducing the overall document feature vector dimension for the random forest proxy labeling model; 2) provide a set of labeling functions that can be leveraged by algorithms like Snorkel-MeTal. We can see in Table~\ref{tab:main-result}, this approach produces two best performing final models.

\subsection{Performance of Proxy Labeling Model}

\begin{table}[h]
	\tbl{Evaluation of the quality of generated proxy labels on the holdout test set.}
	{\begin{tabular}{@{}lccc@{}}
			\toprule
			&  \multicolumn{3}{c}{Apr 2019 to May 2019} \\
			Labeling model & Avg Accu & EM & Wgt $F_1$ \\ \colrule
			Naive Unlabeled & 81.2 & 40.3 & 53.2 \\
			Exp-guided Unlabeled & 82.8 & 46.1 & 60.0 \\
			Exp-guided Snorkel & 11.5 & 2.6 & 42.3 \\  \botrule
	\end{tabular}}\label{tab:labeling-model-result}
\end{table}

We performed additional analysis to gain more insights into the improved performance of LitGen due to proxy labels on the unlabeled papers. Since we do not have access to ground truth labels for the unlabeled papers, we evaluate the performance of the proxy labeling models on the holdout evaluation dataset that we used to evaluate our BiLSTM model. It is notable that the random forest with explanation-guided feature selection (Exp-guided Unlabeled) gives reasonably accurate proxy labels, and is indeed more accurate than the Naive Unlabeled which does not have this feature selection. Moreover because this random forest derived proxy label provides complementary signal, training the original BiLSTM on this additional data leads to additional improvements and give rise to our final LitGen algorithm. We note that popular weak supervision algorithm, Snorkel, performs poorly \anie{with our automatic labeling functions}.


\subsection{Performance by Evidence Types}

\begin{table}[h]
	\tbl{Accuracy of baseline (always guess the majority class), BiLSTM and LitGen model for each evidence type.}
	{\begin{tabular}{@{}lccc@{}}
			\toprule
			Evidence type & Baseline (Majority) & BiLSTM & LitGen \\ \colrule
			Experimental Studies & 63.1 & 85.6 & \textbf{86.7} \\
			Allele Data & 65.7 & 80.4 & \textbf{83.0}  \\
			Segregation Data & 73.8 & 88.8 & 88.8 \\  
			Specificity of Phenotype & 66.0 & 87.0 & \textbf{90.2} \\
			Case Control & 45.8 & 71.2 & \textbf{76.4} \\
			\botrule
	\end{tabular}}\label{tab:accu-evidence-type}
\end{table}

We show the performance of our model on each of the evidence types in Table~\ref{tab:accu-evidence-type}. We can see that one of the most difficult class to predict is the evidence type ``segregation data''. We conjecture that this is because we only used paper abstracts. Most segregation data are mentioned in the actual content of the paper. However, it remains a major challenge for a deep learning system to consume input as long as a full scientific paper. One of the easiest evidence types to learn is ``experimental studies'' because curators mostly look for experimental procedure keywords and most of them are present in the abstracts.




\section{Discussion}

\paragraph{Automatic literature recommendation for variant curation}
We propose a new goal for the field of literature recommendation: automatically generate semantic tags according to VCI evidence types to aid biocurators in filtering papers. We are operating under a low-resource setting where few papers have currently been annotated by experts. However, such annotations are very rich and often contain explanations to justify curator's decisions to submit a paper as evidence. We propose a pipeline that leverages explanations beyond semantic parsing and can be automatically learned by training a Lasso classifier. 


\paragraph{Implication for the curation pipelines}
In the era of implementing genomic medicine, machine assistance is needed for scalability. Human time should be reserved for steps that need true domain expertise and critical interpretation. A feasible model for systematic curation at scale would automate the generation and delivery of gene or variant level information to expert biocurators that can then critique the quality and relevance of the evidence in the context of a specific disease.
This reduces the time it takes to identify evidence of interest that need more in depth human review. 

Our machine learning model for predicting relevant literature by variant and evidence type is well suited for a semi-automated model of curation at scale. Early efforts in automated literature curation have been able to \anie{recommend} papers by \anie{matching} for the variant of interest. The added functionality  \anie{suggests} what type of evidence helps to further streamline curation workflow and efficiency by pre-mapping evidence onto predicted ACMG/AMP criteria. Displaying papers by evidence type also matches the natural organization of curation interfaces such as the VCI, making this an even more feasible tool to implement and have true clinical impact. LitGen is not meant to replace biocurators, but rather to \anie{facilitate} the curation process by prioritizing papers that are more likely to contain particular types of evidence. 

\paragraph{Acknowledgement} This project is supported by NIH ClinGen. J.Z. is also supported by the Chan-Zuckerberg Biohub. We would like to thank S. Plon and A. Milosavljevic for feedback.

\bibliographystyle{ws-procs11x85}
\bibliography{ws-pro-sample}

\end{document}